\let\labelindent\relax

\documentclass[letterpaper, 10 pt, conference]{ieeeconf}  

\IEEEoverridecommandlockouts                              

\usepackage{graphicx}
\usepackage{float}
\usepackage{xcolor}
\usepackage{url}
\usepackage{hyperref}
\usepackage{mathtools}
\usepackage{amsfonts}

\usepackage{listings} 
\usepackage{todonotes}
\usepackage{soul}
\let\labelindent\undefined
\usepackage{enumitem}
\usepackage{bm}

\newcommand{\norm}[1]{\left\lVert #1 \right\rVert}

\title{\LARGE \bf
Bi-Manual Block Assembly via Sim-to-Real Reinforcement Learning
}

\author{Satoshi Kataoka$^{1}$, Youngseog Chung$^{{2}^*}$, {\small Seyed} Kamyar {\small Seyed} Ghasemipour$^{{1}^*}$, \\Pannag Sanketi$^{1}$, Shixiang Shane Gu$^{1}$, Igor Mordatch$^{1}$%
\thanks{$^{*}$Authors contributed equally to this work}%
\thanks{$^{1}$Google Research}%
\thanks{$^{2}$Machine Learning Department, Carnegie Mellon University}%
}

\begin{document}

\maketitle
\thispagestyle{empty}
\pagestyle{empty}

\begin{abstract}

Most
successes in robotic manipulation have been restricted to single-arm gripper robots, whose low dexterity limits the range of solvable tasks to pick-and-place, insertion, and object rearrangement. More complex tasks such as assembly require dual and multi-arm platforms, but 
entail a suite of unique challenges such as bi-arm coordination and collision avoidance, robust grasping, and long-horizon planning.
In this work we investigate the feasibility of training deep reinforcement learning (RL) policies in simulation and transferring them to the real world (Sim2Real) as a generic methodology for obtaining performant controllers for real-world bi-manual robotic manipulation tasks.
As a testbed for bi-manual manipulation, we develop the ``U-Shape Magnetic Block Assembly Task", wherein two robots with parallel grippers must connect 3 magnetic blocks to form a ``U"-shape.
Without a manually-designed controller nor human demonstrations, we demonstrate that with careful Sim2Real considerations,
our policies trained with RL in simulation enable two xArm6 robots to solve the U-shape assembly task with a success rate of above 90\% in simulation, and 50\% on real hardware without any additional real-world fine-tuning.
Through careful ablations, we highlight how each component of the system is critical for such simple and successful policy learning and transfer, including task specification, learning algorithm, direct joint-space control, behavior constraints, perception and actuation noises, action delays and action interpolation. 
Our results present a significant step forward for bi-arm capability on real hardware,
and we hope our system can inspire future research on deep RL and Sim2Real transfer of bi-manual policies, drastically scaling up the capability of real-world robot manipulators. The accompanying project webpage and videos can be found at: \href{https://sites.google.com/view/u-shape-block-assembly}{sites.google.com/view/u-shape-block-assembly}.



\end{abstract}

\section{Introduction}



Robot arms have the immense potential to perform a wide variety of useful manipulation tasks, and approaches have already made their way into many applications such as manufacturing and warehouse pick-and-place operations~\cite{eppner2016lessons}. However, most successful cases are confined to the single-arm setting~\cite{pinto2016,asyncnaf2017,qtopt2018,levine2018,insertion2020,tossingbot2020}.
Dual and multi-arm platforms enable robots to accomplish a much wider variety of complex tasks, such as cooking and construction, due to additional dexterity and effective workspace (e.g. richer grasp wrench space~\cite{grasping2016}).
Furthermore, the use of multi-arm robots can simplify tasks that can be accomplished by single-arm robots, by alleviating the need for extremely dexterous in-hand manipulation behaviors~\cite{rubik2019,chen2021system}.

\begin{figure}[t]
    \centering
    \includegraphics[scale=0.62,bb=0 0 490 290]{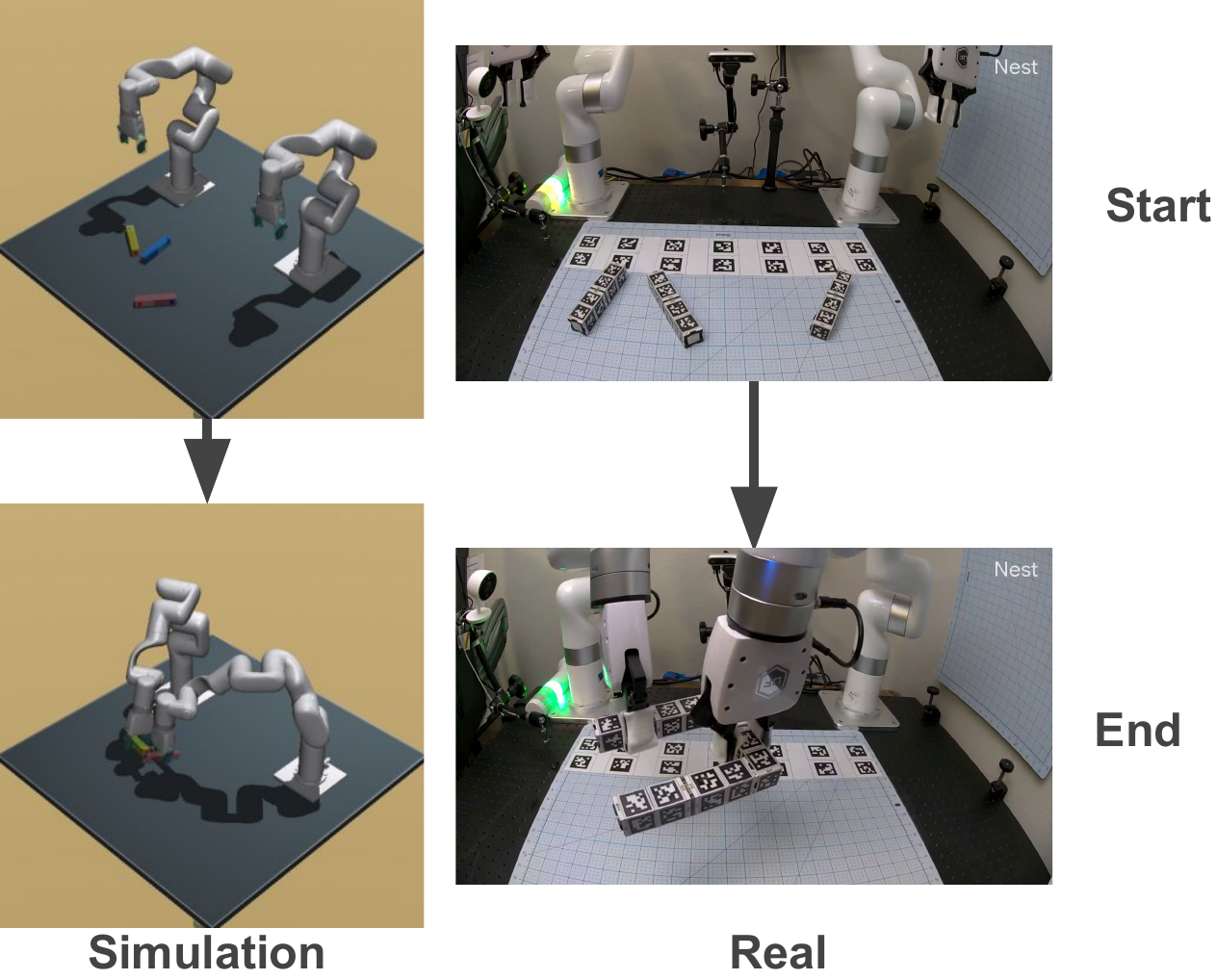}

    \caption{
        \scriptsize{Visualization of simulated and real-world U-Shape Assembly Tasks.}
    }
    \label{fig:three_block_task}
\vspace{-6mm}
\end{figure}

Despite their inherent advantages, introducing multiple arms into robotic systems brings about a host of unique challenges to the control problem. Designing controllers for bi-manual systems in end-effector or task space is often ineffective as one needs to consider complex inter-arm collision avoidance and coordination behaviors, especially for tasks such as assembly which require long-horizon, intricate, and close interactions among arms and objects. Motion planning algorithms~\cite{lavalle1998rapidly,kavraki1996probabilistic} are intractable for closed loop real-time multi-arm manipulation settings due to their unfavorable scaling as the degrees of freedom in systems increase. On the other hand, open loop controllers are also ineffective in multi-arm setups, as they are more likely to experience drift from the intended plan due to noise, asynchronous control delays, and amplification of coordination errors. Thus, in current robotics literature, most examples of multi-arm manipulation systems rely on planning with parameterized motion primitives \cite{amadio2019}, imitation learning \cite{peg2022}, or reinforcement learning with parameterized sub-motions~\cite{luck2017,chitnis2020efficient}.

In contrast to the above approaches, the use of reinforcement learning (RL) for joint-space control of bimanual robots is significantly underexplored. In particular, with the use of deep learning, RL methods have many significant advantages: neural network policies can be seamlessly integrated with perception modules that are themselves often reliant on deep learning techniques~\cite{pinto2016,qtopt2018,levine2018,saycan2022}; neural network policies can be trained in multi-task setups and effectively share acquired knowledge amongst tasks~\cite{metaworld2020,lee2022multi,reed2022generalist}; the output of an RL system is a policy that can be executed with tight real-time real-world constraints~\cite{isim2real2022}. Thus, in this work we investigate the feasibility of training deep RL policies in simulation and transferring them to the real world (Sim2Real), as a generic methodology for obtaining performant controllers for real-world bi-manual robotic manipulation tasks.

To study the efficacy of simulated deep RL and Sim2Real transfer, we design the \textit{``U-Shape Magnetic Block Assembly Task"} as a testbed for studying key challenges of bi-manual manipulation in the real world. In this task, two robot arms equipped with parallel grippers must connect 3 magnetic blocks in order to form a "U" shape. Despite its minimalism, magnetic block assembly is substantially more challenging than prior bi-manual policy learning tasks on real robots~\cite{luck2017,amadio2019,peg2022}, and preserves many of the core challenges of multi-arm manipulation such as coordination, precision, geometric reasoning and collision avoidance (between robots, blocks, and ground), and real-time execution constraints.
Furthermore, solutions are required to be closed-loop as blocks can move, slip, and drop. Block assembly can also be readily extended by increasing the number and diversity of available blocks, and replacing magnets with altenative attachment mechanisms to enable the construction of more complex blueprints.

In this work, we demonstrate that deep RL and Sim2Real transfer lead to an effective recipe for solving the ``U-shape block assembly task". The key contributions of our work are as follows:

\begin{itemize}[leftmargin=10pt]
    \item We develop the ``U-Shape Magnetic Block Assembly Task" as a testbed for real-world bi-manual manipulation, and detail each component of the system design in simulation and the real world.
    \item To the best of our knowledge, we demonstrate for the first time that joint-based control through deep reinforcement learning is an effective approach for obtaining controllers for a real-world bi-manual manipulation task.
    \item Critically, we demonstrate that by taking the appropriate measures, RL policies trained in simulation can be transferred to the real-world without any fine-tuning. We present detailed ablations identifying key components for effective Sim2Real transfer of RL policies trained in simulation.
\end{itemize}

 \begin{figure*}[t]
            \begin{minipage}{\textwidth}
                \centering
                \includegraphics[scale=0.5]{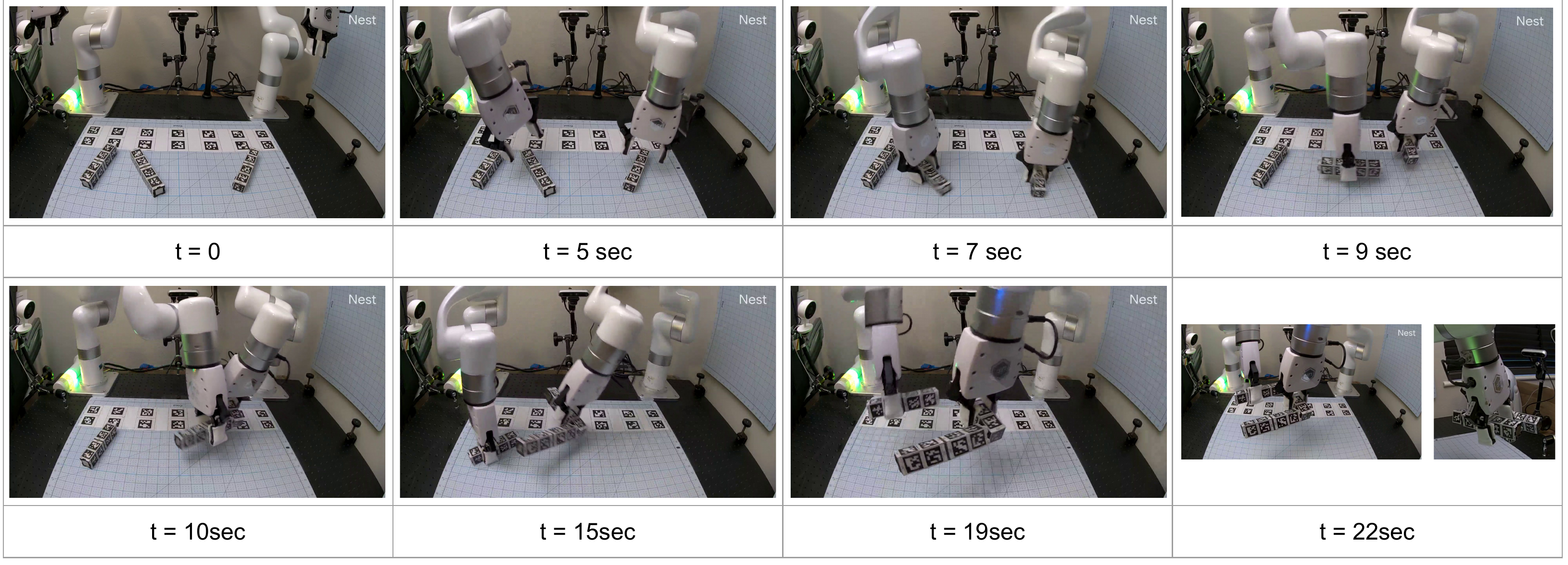}
                \vspace{-3mm}
                \caption{
                    \small{
                        Snapshots of U-Shape Block Assembly Task Execution in Real World.
                    }
                }
                \label{fig:real_world_demo}
                \vspace{-4mm}
            \end{minipage}
        \end{figure*}


Our results present a significant step forward for bi-arm capability on real hardware.
Since our policy is learned through RL with many hours of simulated experience, it also exhibits a series of emergent properties such as intentional ungrasping,
robust retries, and intentional slow-downs. While most learning-based robotic research focuses on enabling perceptual~\cite{pinto2016,qtopt2018,levine2018} and semantic~\cite{saycan2022} generalization, we hope our work could inspire the community to look more into fundamentally enhancing the dexterity and effective workspace of robots in the real world through machine learning.

\section{Related Works}

\textbf{Deep RL and Sim2Real for Robot Learning:}
In recent years, deep RL has led to many successes in the field of robotics \cite{asyncnaf2017, qtopt2018, rubik2019}. However, training neural network policies through RL requires millions of environment interactions for even the simplest of tasks \cite{dmcontrol2018} \footnote{Indeed, as we will describe below, training policies for our ``U-shape Block Assembly Task" requires $\sim$1 billion environment steps, which amounts to about 3000 days of robot operation time.}. This, in addition to concerns of hardware cost and safety, makes RL policies infeasible to train directly in real-world setups. Thus, in this work, we focus on training bi-manual manipulation policies in a simulated setup.
    
Unfortunately, policies trained in simulation rarely transfer well to real-world systems due to model discrepancies and noise in real-world state-estimation. This problem -- often referred to as the ``Sim2Real'' problem -- has led to a significant body of work in recent literature \cite{sim2real2017,sim2real2018,rubik2019,isim2real2022}. Our policy learns without human demonstrations and transfers successfully to the real robots without on-robot fine-tuning.

\textbf{Bi-Manual Robot Learning:} In contrast to the large body of work and benchmarks for single-arm manipulation~\cite{her2017,asyncnaf2017,qtopt2018,metaworld2020,rlbench2020} and dexterous in-hand manipulation~\cite{kumar2016,rubiks2020,baoding2020,robel2020}, those for bi-manual are comparatively limited due to various additional difficulties, including higher action dimensions and larger workspace. Policy learning on real robots is studied in \cite{luck2017} for bi-manual pick up, \cite{amadio2019} for towel folding, and \cite{peg2022} for bi-manual peg insertion (using human demonstrations). \cite{robosuite2020} developed a rich simulation benchmark including bi-manual pickup and peg insertion tasks. \cite{bidex2022} designed a suite of tasks with two floating five-finger-hand manipulators, but despite high dexterity of the tasks, these are only within simulation and the learned behaviors look unnatural for real robot deployment. Our results present a substantial advancement compared to these prior bi-manual tasks, as we require long-horizon planning with multi-step grasping and complex collision avoidance to account for close and intricate bi-arm interaction. 
    
\textbf{Assembly and Construction:}
Most prior learning-based assembly results were only in simulators, often with unrealistic physics simplifications to improve learning through RL. For example, \cite{bapst2019structured} studied 2D simulated task-oriented construction environments, but many realistic and difficult details of assembly are abstracted away, as the agent has the ability to directly actuate a block of choice anywhere in the scene and weld blocks via an explicit action.
\cite{blockassemble2022} studied 3D simulated block assembly environments with similar direct block actuation and achieved combinatorial generalization, assembling target blueprints consisting of up to 16 blocks using a graph neural network policy. \cite{lee2019ikea,lee2021adversarial} also introduce a 3D simulated assembly environment for furniture design from a blueprint, but with limited success in policy generalization. In contrast, our environment contains all of the necessary features for real-world execution (e.g. physical blocks, full robot arm with gripper model, noises and actuation constraints). Also, we successfully executed real-time joint-space control by our agent in the real-world to construct a U-shaped structure where three magnetic blocks are connected.
\cite{suarez2018can} studied assembly of a single chair with real-world bi-manual robots using offline planning methods. In contrast, we use a neural network policy which generates online actions. 
As we will discuss in Section~\ref{sec:emergent_behaviors}, the neural network policy is even able to produce robust emergent behavior, e.g. retries of picking up after accidentally dropping the blocks.


\section{Task Setup}
\label{sec:task_setup}
Our goal is to study how to obtain controllers for real-world bi-manual manipulation systems. To this end, we design a magnetic block assembly task as a testbed.
    \subsection{Task Description}
        \label{sec:tasks}
        We study the \textit{U-Shape Block Assembly} task (Figure~\ref{fig:three_block_task}) where three blocks with magnetic connection points (Figure~\ref{fig:icra_blocks}) are placed on the ground, and two robotic arms must attach the three blocks to form a U-shape.
        While our long-term objective is to enable assembling any number of blocks, we design this task as a minimal configuration that stresses bi-arm coordination in the real world.
        Despite its minimalism, increasing the number of magnetic blocks can support the creation of arbitrarily complex composed structures, which can lead to many intriguing avenues of future research.
        In our U-shape block assembly task, each trial continues indefinitely until 1) system identifies all three blocks are connected, 2) robots get faulted, 3) robots are manually stopped to prevent breakages (in real only), 4) 50-second timelimit passes (in sim only).
        Determination of success is different between sim and real, according to the different information available in each environment. In simulation, we programmatically verify the connections of the correct magnet pairs to evaluate success. In the real world, at the end of each trial, a human operator verifies that the correct magnetic connections are successfully made. If the robot arms run into unsafe behavior in the real world, the human operator terminates the trial and marks it as a failure.
     

    \subsection{Hardware Platform}
\begin{figure}[t]
            \begin{minipage}{0.22\textwidth}
                \centering
                \includegraphics[scale=.16]{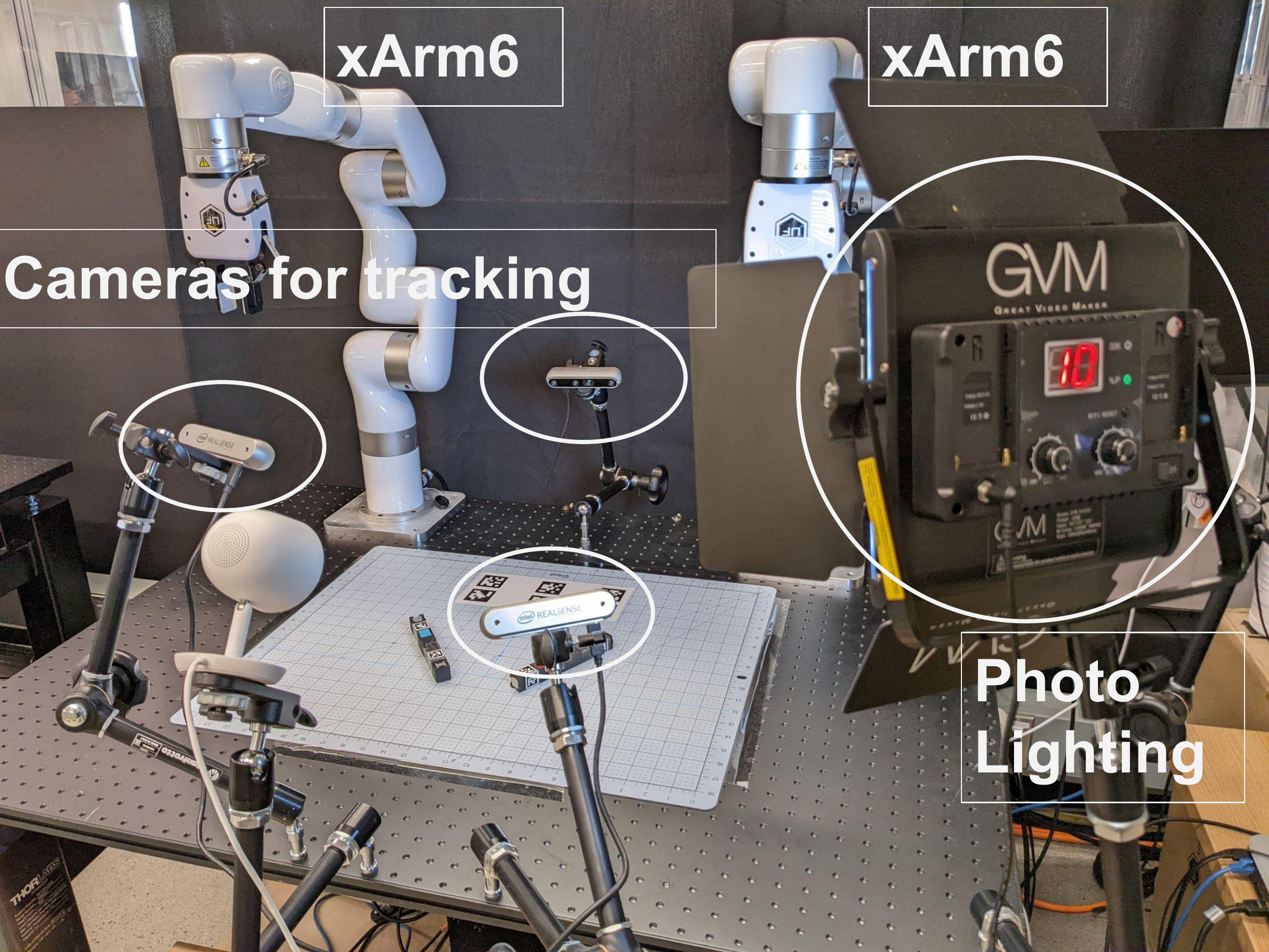}
                \caption{Robot Setup.}
                \label{fig:icra_cameras}
            \end{minipage}        
            \begin{minipage}{0.23\textwidth}
                \centering
                \includegraphics[scale=.16]{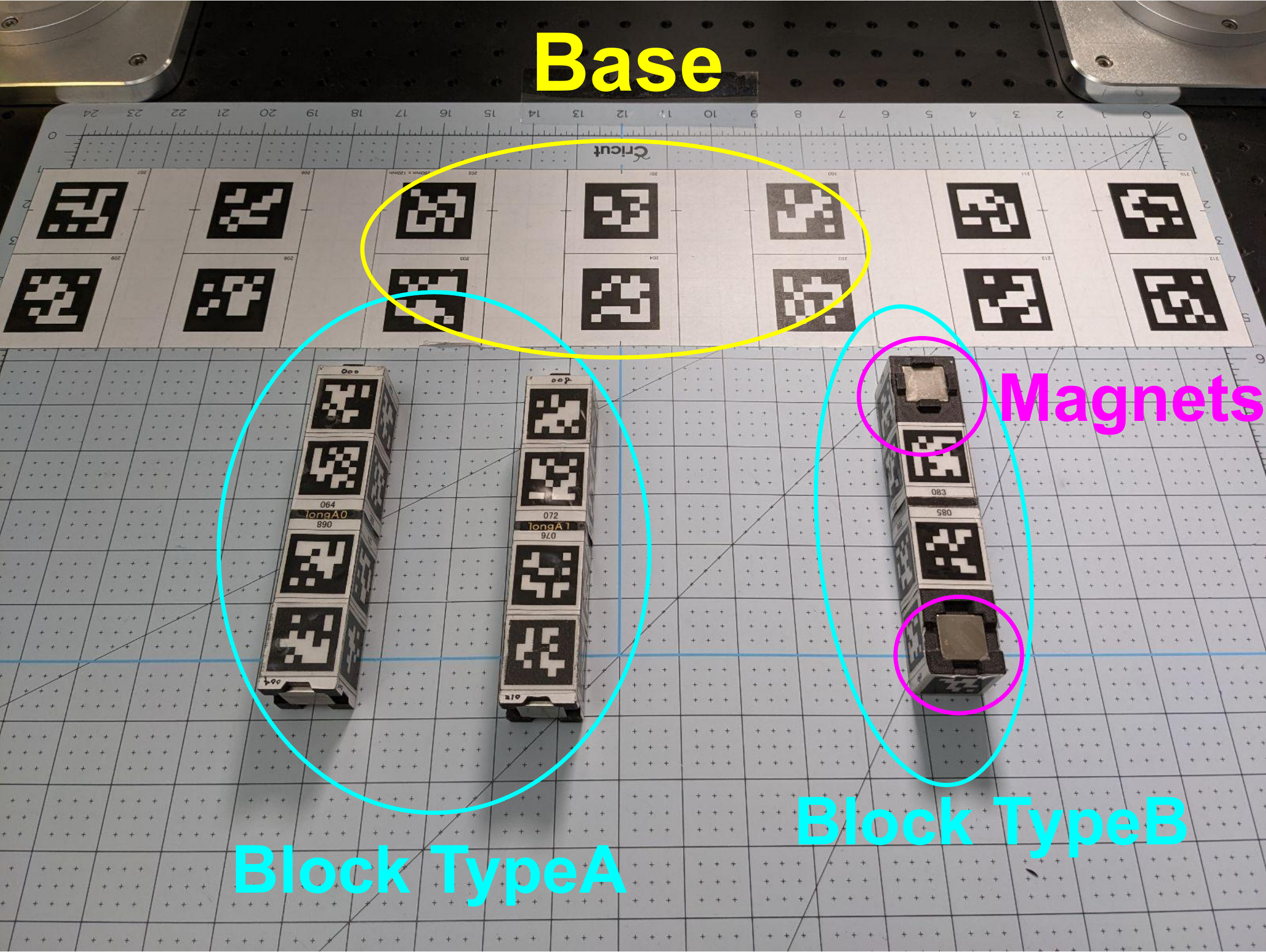}
                \caption{
                        Magnetic blocks.
                }
                \label{fig:icra_blocks}
            \end{minipage}\hfill
            \vspace{-6mm}
        \end{figure}

        We now describe the real-world setup for the block assembly task.
        We construct a physical work cell (Figure~\ref{fig:icra_cameras}) which consists of two robotics arms (UFACTORY xArm6 \cite{xarm6}), three cameras (Intel RealSense D455 \cite{d455}), and three cuboid blocks with magnetic connection points (Fig~\ref{fig:icra_blocks}). To estimate the position and orientation of the blocks, we use the AprilTag \cite{apriltag} tracking library. All blocks have AprilTag markers attached to their surfaces. We also put AprilTag markers on the workcell floor to track relative positions between the base and blocks. Blocks are printed with a 3D printer using PLA, and consist of two TypeA and one TypeB blocks (Figure~\ref{fig:icra_blocks}):
        \begin{itemize}[leftmargin=*]
            \item \textbf{TypeA ($\bm{\times2}$)}: Size is 30mm $\times$ 30mm $\times$ 150mm. Two magnets (18mm $\times$ 18mm) of opposite polarizations are embedded onto each of the two square sides.
            \item \textbf{TypeB ($\bm{\times1}$)} Size is also 30mm $\times$ 30mm $\times$ 150mm. Two same magnets of opposite polarization are embedded onto one long side of the block.
        \end{itemize}
            
   For tracking, we use three Intel RealSense D455 \cite{d455} cameras. While the cameras can provide depth images as well as RGB, we only utilize the RGB images at a resolution of 1280 $\times$ 800 at 30fps. The produced images are used for the AprilTag tracking system \cite{apriltag} and visualization of experiments. Figure \ref{fig:icra_cameras} shows the camera placements.

    \subsection{Direct Joint-Space Control}
        In this work we have decided to train policies for real-time (4Hz) joint velocity control of the robot arms. As opposed to Cartesian control of end-effector positions, joint-space control not only alleviates the need for inverse kinematics (IK),
        but enables policies to fully utilize the joint-space to learn optimal bi-manual behaviors that avoid collisions between the robot arms while executing the desired task.
        The policy produces joint \textit{velocity} controls at 4Hz, and we convert the joint velocity controls to 100Hz joint \textit{position} commands which are directly sent to the xArm6 controller.

    \subsection{Simulated Platform}
        We construct the simulated counterpart to our real-world setup using the Mujoco simulator~\cite{mujoco2012}, which we refer to as the \textit{magnetic assembly environment}. This environment contains the same set of 3 cuboid blocks of the two different types (TypeA and TypeB, Fig~\ref{fig:three_block_task}), where positive and negative magnets are rendered as red and blue respectively.
        Positive and negative magnets ``snap" together when they are within 1cm and disconnect when adequate pulling force is applied.
        In each training episode, the simulated environment starts with the blocks randomly dispersed on the ground, and the robots are reset to predefined initial poses as shown in Figure \ref{fig:three_block_task}, top.
        Episodes are 200 environment steps long, translating to a length of 50 seconds in the real world.
        
        At each stage of the assembly process, we assign each gripper $g_i$ to a particular block $b_i$ that it must manipulate, and the two arms must connect the two corresponding blocks using the correct magnet pairs. The assignments are scripted and dependent on which magnets are currently connected.
        We made design choices in the RL reward function (a function of the current state, $s_t$) to elicit the connecting behavior, and the following three additive terms (each weighted by $w$'
        s) were effective for this goal.
        The first reward term is the sum of distances between each $i^{\text{th}}$ gripper's position
          $p_{g_i}\in\mathbb{R}^3$ and its target block's center-of-mass position $p_{b_i}\in\mathbb{R}^3$:
        \vspace{-2mm}
        \begin{equation}
            r_{grasp}(s_t) = -w_g \sum_{i=0}^1\norm{ p_{g_i} - p_{b_i}}_2.
            \vspace{-2mm}
        \end{equation}
        The second and third reward terms are for aligning two correct magnet pairs. For a desired connection $j$, let the positive and negative magnet positions be denoted by $p_j, q_j \in \mathbb{R}^3$, and their rotation matrices denoted by $u_j, v_j \in \mathbb{R}^{3\times3}$ respectively: 
        \vspace{-2mm}
        \begin{equation}
            r_{mpos}(s_{t}) = -w_{mp} \sum_{j=0}^1 \norm{ p_{j}-q_{j} }_2 
            \vspace{-7mm}
        \end{equation}
         \begin{equation}   
            r_{mrot}(s_{t}) = -w_{mr} \sum_{j=0}^1(1.0 - \arccos(u_j[2] \cdot v_j[2])).
            \vspace{-2mm}
        \end{equation}

        
    \subsection{Environment Observation and Action}
        When designing observations to provide to agents, we have taken the effort to ensure the same observations are available in both simulation and the real world.
        \paragraph{Block Observations}
        We use very minimal block observations, namely the global position and orientation (represented as a 3x3 rotation matrix) for all blocks. In the real world environment, we simply transform tracking results from our AprilTag tracking system for the blocks to match our simulator's coordinate system, without using any additional statistical estimation filtering, and directly feed them as input to the policy.
        \paragraph{Robot Observations}
        For each robotic arm, we include as observations: current joint positions, current end-effector positions, target joint positions in the previous timestep, and the extent of gripper opening width
        in the previous timestep. In the real world environment, these states are obtained from the robot controllers.
        
        To improve tolerance to noises incorporated into the simulator (Section~\ref{sec:perception_and_actuation_noise}), we form the complete environment observations by concatenating the observations from current and previous timesteps.
    
        \paragraph{Robot Action}
        Our agent's action consists of joint velocities and gripper opening width for two arms.
        
        

\section{Learning Setup}
    \label{sec:sim}
    To solve the U-shape block assembly task, we use RL to learn robust control policies in simulation,
    and transfer learned policies to the real-world without any additional fine-tuning.
    First, we describe the learning setup in simulation.

    \subsection{Policy Network Architecture}
        Observations are fed to a neural network policy which consists of a 4-layer MLP, with 1024 dimensional hidden layers and swish non-linearities. The output of the policy network consists of two vectors representing the mean and standard deviations for a diagonal multivariate Gaussian distribution. A sample is then squashed by an elementwise Tanh function, and used as the control action.

    \subsection{Algorithm}
       \paragraph{Distributed PPO} We train our RL policies using Proximal Policy Optimization (PPO) \cite{schulman2017proximal} with Generalized Advantage Estimation (GAE) ~\cite{schulman2015high}, and follow the practical PPO training advice of \cite{andrychowicz2020matters}. As will be shown below, a key ingredient in enabling the training of our bi-manual object manipulation agents is the scale of training.
       Our distributed PPO agents are implemented using Jax~\cite{jax2018github}, Haiku~\cite{haiku2020github}, and Acme~\cite{hoffman2020acme}. Agents are trained for 2-3 billion environment steps, using 1 Nvidia V100 GPU for training, and 4000 preemptible CPUs for generating rollouts in the environment, capable of generating 1 billion environment steps in 8 hours (35,000 steps per sec). 
       
       \paragraph{Distributed SAC (ablation)} We also train policies with a distributed Soft Actor-Critic (SAC)~\cite{sac2018} implementation provided by Acme~\cite{hoffman2020acme}.
        Using identical compute infrastructure, and despite training for over 72 hours, SAC was unable to solve the U-shape assembly task, and had a success rate of 0\%.
        In contrast to the successful application of SAC to standard deep RL control benchmarks~\cite{dmcontrol2018} and single-arm manipulation tasks~\cite{metaworld2020}, our results emphasize the difficulty of our U-shape assembly task, which requires both high precision contact and long-horizon coordination, and is not solvable by the aggressive exploration behavior of SAC.
\section{Overcoming the Sim-To-Real Gap}
\label{sec:sim2real}


Many of the choices in Task Setup (Section~\ref{sec:task_setup}) and Learning Setup (Section~\ref{sec:sim}) have been specifically made for successful simulation-to-real transfer, without any additional finetuning of RL policies in the real world.
In this section we present additional design details for overcoming the simulation-to-real gap, which have proven to be critical in our experimental ablations (Section~\ref{sec:real_results}).
    \subsection{Behavior Constraints}
    \label{sec:behavior_constraints}
        \paragraph{Joint Velocity and Acceleration Limits}
        To enable joint-space policies to produce feasible joint position targets in the real world, it is important to reproduce strict constraints for joint velocity and acceleration limits in simulation. To this end, the policy outputs -- which are target joint velocities for the robotic arms -- are clipped in order to satisfy the joint acceleration limits.

        \paragraph{Applied Force Limit}
        \label{sec:applied_force_limit}
        Excessive forces applied to the robots can cause robots to fault in the real world.
        Additionally, high applied forces may easily lead to object breakages in real world, even if the applied forces are within the robots' configuration limits. To encourage policies to learn less aggressive behaviors and avoid excessive forces, in each simulation step we calculate applied forces for all contacts and apply negative rewards for violations of force limits. We have two types of the applied force limitations. The first one is a general applied force limit which is used for all contacts. This limitation helps the policy avoid causing unrealistic forces between all objects. The second limit is an applied force limit between the blocks and the floor. This limitation helps the robots avoid pushing gripped objects to the ground, which can easily lead to robots' faulting due to joint load limits.
        In the Mujoco simulator, information regarding applied forces can be obtained through a ``constraint force" function. Although force limit is an important factor for successful Sim2Real transfer, limiting applied force significantly increases task difficulty for RL training. We show the training performance across different force limits in simulation in Section~\ref{sec:applied_force_limt_result}.
    
    \subsection{Perception and Actuation Noise}
    \label{sec:perception_and_actuation_noise}
        In the real world, perception data from sensors are inevitably noisy. To address this problem, instead of using statistical filters (e.g. Kalman filter) at runtime, we opted for incorporating noise into our simulation. Specifically, we applied a zero-mean diagonal Gaussian noise (noise scale estimated from real-world data) to both the positions and the orientations of the blocks. Incorporating noise directly into our simulation environment prevents the policy from learning highly agile behaviors that take advantage of idiosyncracies of simulator physics, and results in more conservative policy behaviors that are robust to observation noise.
        Additionally, this allows us to remove an extra abstraction layer for filtering sensor data in the real world,
        which not only contributes to the simplicity of the system, but also reduces the gap between the simulation and real world software stacks.  We show results of real world demonstration with and without noise in real world result Section~\ref{sec:real_results}.
        
    \subsection{Action Delay and Interpolation}
\begin{figure}[t]
            \centering
            \includegraphics[scale=.165]{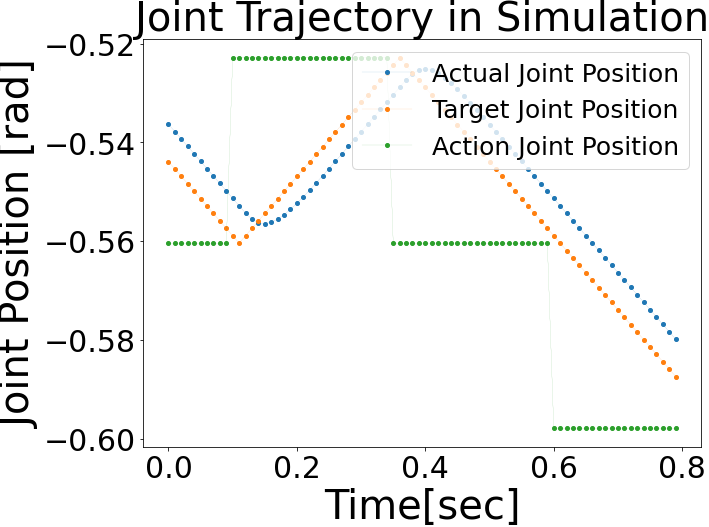}
            \includegraphics[scale=.165]{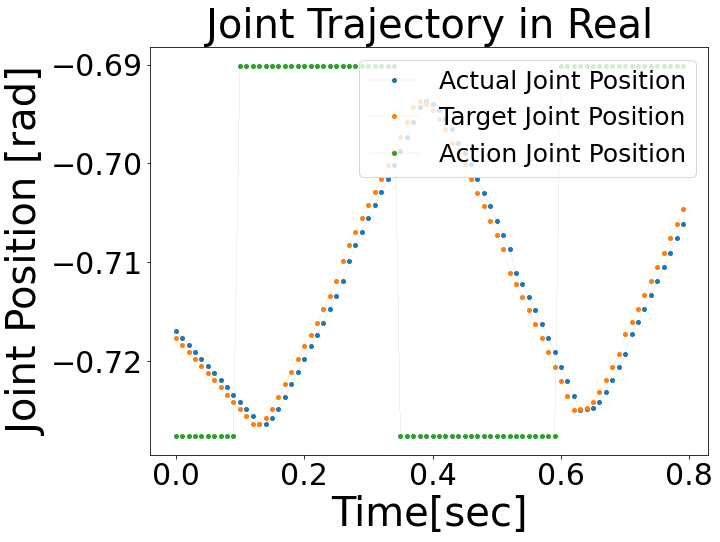}
            \caption{
                \small{
                    Joint Trajectory with Delayed and Interpolated Actions
                }
            }
            \label{fig:action_delay_and_interpolation_joint_graph}
            \vspace{-5mm}
        \end{figure}

        In zero-shot Sim2Real transfer, the difference of joint behavior between sim and real is a critical gap which prevents the policy from reproducing the expected behaviors in the real world. To address this issue, we characterize the joint behavior by defining 1) the delay between application of joint action and joint action fulfillment, and 2) the start time and duration of action interpolation (policies run at 4Hz and actions are interpolated for 100Hz control). We apply these two parameters in both sim and real environments in order to reproduce the same joint behavior in sim and real. Figure~\ref{fig:action_delay_and_interpolation_joint_graph} shows how the produced action is interpolated and delayed before sending commands to the robot in both sim and real. These graphs demonstrate how the joint position actions are converted to joint position command and fulfilled on the robots. Note that the joint position action is the sum of the last target joint position and the policy's velocity action output.
        We show the effect of modeling these parameters in the real world results in Section~\ref{sec:real_results}.


\section{Experiments And Results}

        
        The goal of our experiments is to validate that learning and transfer of bi-arm policy for assembly is feasible, and to evaluate which pieces are critical through ablation studies.
        
        In summary, our learning setup can consistently produce a PPO policy with $\geq$95\% success rate in 3B train steps, 24 hours in simulation,
        which retains 50\% success rate when zero-shot transferred to the real world. Figure \ref{fig:real_world_demo} shows snapshots of the behaviors in the real world execution, generated by the agent trained on the U-shape block assembly environment. In the snapshots, the two robotic arms accurately grip the blocks and move the blocks to the connecting position, open left gripper to let the blocks connect via the magnets, grip the 3rd block, move the 3rd block to the 2block structure, then open the gripper to let the 3rd block connect. In the following sections, we show ablations in both simulation-based learning and Sim2Real design choices which were crucial for these results.
        
    \subsection{Simulation RL Results}
        \label{sec:sim_results}
    
        We evaluate the performance of a policy on the U-shape block assembly task in simulation by computing the average success rate over 40 episodes, with ablations on different network sizes,  applied force limits and noises (Section \ref{sec:behavior_constraints}). Given the high computation cost of large-scale distributed training (using 4000 CPUs), we run at most 2 seeds for each experiment and choose the better one for evaluation.

        \textbf{Ablation: Effect of Network Width and Depth}
        To check the effect of policy network size on training progress, we trained the policy with different network size configurations (256-2048 as MLP hidden units, and 1-6 as number of MLP layers).
        For each setting of network depth and width, we measure the number of train steps required to reach 90\% success rate in simulation.
        The results indicate that training is optimized when the number of hidden units is between 512 and 1024, and when the number of network layers is between 2 and 4.
        Given these results, to ensure a large network capacity, we configure our policy networks to have 4 layers of 1024 hidden units.

        \textbf{Ablation: Applied Force Limit}
        \label{sec:applied_force_limt_result}
        Excessive applied force in the real world poses safety risks, thus the policy must learn behaviors which are not too forceful, and to achieve this, we place a limit on applied force in simulation. However, this increases the task difficulty.
        In Table~\ref{tab:block_to_floor_force_limit}, we show the effect of force limit on training speed. As expected, tightening the limit (lower value) makes training much slower, and limits which are too low can even prevent the policy from learning the task.
      
        In determining the block to floor force limit, we find that when policies learned with a limit of 3.0 or higher are deployed in the real world, they frequency cause abnormal electric current fault. Therefore, we apply 2.5 as the block to floor applied force limit.
        Likewise, we found 5.0 to be an adequate value for the general applied force limit.
        (Note: the unit for force limit is defined by Mujoco~\cite{mujoco2012}.)
        \noindent
        \begin{minipage}{0.50\textwidth}
            \centering
            \begin{table}[H]
                \scriptsize
                \setlength\tabcolsep{10pt}
                \centering
                \begin{tabular}{ |c|c|c|c|c|c|}
                    \hline
                    \multicolumn{1}{|c||}{} & \multicolumn{5}{c|}{Block To Floor Applied Force Limit} \\ 
                    \multicolumn{1}{|c||}{} & 1.0 & 1.5 & 2.5 & 3.75 & 5.0 \\ 
                    \cline{2-5} \hline
                    \multicolumn{1}{|c||}{Steps} & N/A & 1420 & 1220 & 1030 & 840 \\ \hline
                    \multicolumn{1}{|c||}{} & \multicolumn{5}{c|}{General Applied Force Limit} \\
                    \multicolumn{1}{|c||}{} & 1.0 & 2.0 & 3.0 & 4.0 & 5.0 \\ 
                    \cline{2-5} \hline
                    \multicolumn{1}{|c||}{Steps} & N/A & N/A & N/A & 2600 & 840 \\ \hline
                \end{tabular}
                \caption{\scriptsize{Effect of applied force limit: training steps (in millions) to reach $90\%$ success rate on U-shape block assembly in simulation.  N/A implies $<10\%$ success after 3 billion steps.}}
                \label{tab:block_to_floor_force_limit}
                \vspace{-0mm}
            \end{table}
        \end{minipage} 
        
        \textbf{Ablation: Effect of Noise in Simulation}
        Similar to applied force limits, modeling noise in simulation is a necessary component for successful policy transfer in the real world, but it also significantly decreases training speed in simulation. Based on the level of observation noise in the real world, we apply in simulation a zero-mean Gaussian noise distribution with scale $=1$cm. This increases training time required to achieve $90\%$ success rate by $\sim1.8$ times. However, as we will see in the next section, modeling noise in simulation is absolutely critical for real world transfer.
        
        These results highlight a key point in developing a functional Sim2Real system for challenging tasks: there can be a trade-off between learning difficulty in simulation and Sim2Real transfer success in the real world, hence these two must be carefully balanced.


    \subsection{Real-World Transfer Results}
        \label{sec:real_results}
        \textbf{Ablation: Effect of Noise and Action Interpolation.}
        We demonstrate the effect of modeling observation noise and interpolating across delayed actions during training on real world performance. 
        In Table~\ref{tab:real_world_results}, we train policies with and without noise and action interpolation applied in simulation, and evaluate its success rate reaching intermediate checkpoints, or sub-tasks, during the U-shape block assembly task. \textit{2 Block Pickup} denotes gripping the first two assembly blocks, \textit{2block Connection} denotes successfully connecting the first two blocks, \textit{3rd Block Pickup} denotes successfully gripping the third block, and \textit{3block Connection} denotes successful U-shape block assembly.
        While applying noise or action interpolation individually improved performance over not applying both, applying both was absolutely beneficial across all sub-tasks.

    \noindent
        \begin{minipage}{0.5\textwidth}
            \centering
            \begin{table}[H]
                \scriptsize
                \setlength\tabcolsep{4pt}
                \centering
                \begin{tabular}{ |c|c|c|c|c|c|}
                    \hline
                    \multicolumn{2}{|c||}{Setting} & \multicolumn{1}{c|}{2 Block} & \multicolumn{1}{c|}{2block} & \multicolumn{1}{c|}{3rd Block} & \multicolumn{1}{c|}{3block}  \\ 
                    \cline{1-2}
                    \multicolumn{1}{|c|}{Noise} & \multicolumn{1}{c||}{Action Interp.} & \multicolumn{1}{c|}{Pickup} & \multicolumn{1}{c|}{Connection} & \multicolumn{1}{c|}{Pickup} & \multicolumn{1}{c|}{Connection}\\ \hline
                    \multicolumn{1}{|c|}{\textbf{YES}} & \multicolumn{1}{c||}{\textbf{YES}} & 95\% & 90\% & 70\% & 50\% \\  \hline 
                    \multicolumn{1}{|c|}{\textbf{YES}} & \multicolumn{1}{c||}{\textbf{NO}} & 30\% & 30\% & 10\% & 0\% \\  \hline 
                    \multicolumn{1}{|c|}{\textbf{NO}} & \multicolumn{1}{c||}{\textbf{YES}} & 40\% & 30\% & 10\% & 10\% \\  \hline 
                    \multicolumn{1}{|c|}{\textbf{NO}} & \multicolumn{1}{c||}{\textbf{NO}} & 10\% & 10\% & 0\% & 0\% \\  \hline 
                \end{tabular}
                \caption{\scriptsize{Effect of noise and action interpolation ("Interp.") on real-world success rate (across 20 trials each).}}
                \vspace{-3mm}
                \label{tab:real_world_results}
            \end{table}
        \end{minipage}
    \subsection{Emergent Behaviors}
    \label{sec:emergent_behaviors}
     Since the policy is optimized using RL in simulation with initial state randomization and exploration noises, it has acquired unintended but intelligent emergent behaviors. These behaviors are shown in the supplementary videos.
    \begin{itemize}[leftmargin=*]
        \item \textbf{Intentional Ungrasp}: Sometimes due to imperfect grasp, the arms open the gripper and regrasp to enable the success of downstream connection.
      
        \item \textbf{Retries}: The arm sometimes drops the block due to imperfect grasp or loss of block information from occlusion, but can re-grasp and often succeed.
      
        \item \textbf{Change-of-pace}:  Sometimes while connecting, the arms appear to slow down to carefully align and bring the blocks closer where accuracy is required.
    
        
    \end{itemize}
      
  \subsection{Future Work}
    There are numerous promising future work to be done from our minimal task and system.
      \begin{itemize}[leftmargin=*]
          \item \textbf{More blueprints and blocks}: While we focused on 3-block U-shape assembly to test out the critical components of a system in depth, we can readily scale and extend the pipeline to more blocks and intricate blueprints.
          \item \textbf{From perception}: We used AprilTags for state tracking, but for simpler real-world deployment, we would ideally prefer raw vision as input.
      \end{itemize}

\section{Conclusion}

Bi-hand manipulation is taken for granted in humans, but it has been a long-standing challenge for real-world robotic systems due to planning and deployment difficulties. In this work, we present a system for enabling bi-manual assembly of multiple magnetic blocks, with the policy learned via RL in simulation without human demonstrations and then transferred zero-shot to the real world. We detail task and algorithm designs, and provide a list of generic techniques, such as behavior constraints and action interpolation, that could help facilitate other Sim2Real manipulation tasks with larger workspace and multi-arm coordination. We hope our work could serve as a proof-of-concept and inspire a series of future robot learning work to venture into more challenging and dexterous bi-manual manipulation tasks, such as assembly, tool construction, tool use, cloth folding, and cooking, and to deploy such systems in the real world. 




\end{document}